\documentclass[sigconf]{acmart}
\usepackage{amsfonts,amsmath,amsthm,mathrsfs,mathtools,upgreek}

\usepackage{tabularx}
\usepackage{booktabs}
\usepackage[mode=buildnew]{standalone}
\usepackage{hyperref}

\usepackage{algorithm}
\usepackage{algorithmic}

\usepackage{multicol}

\usepackage{natbib}
\usepackage{tikz}
\usetikzlibrary{arrows.meta}
\usepackage{listofitems} 
\tikzstyle{mynode}=[thick,draw=blue,fill=blue!20,circle,minimum size=0.5]
\usepackage{enumitem}
\newcommand{\defeq}{\vcentcolon=}

\newtheorem{rem}{Remark}

\newtheorem{defn}{Definition}
\newtheorem{problem}{Problem}

\PassOptionsToPackage{dvipsnames}{xcolor}
\definecolor{LightGray}{gray}{0.9}

\usepackage{listings}

\definecolor{codegreen}{rgb}{0,0.6,0}
\definecolor{codegray}{rgb}{0.5,0.5,0.5}
\definecolor{codepurple}{rgb}{0.58,0,0.82}
\definecolor{backcolour}{rgb}{0.95,0.95,0.92}

\lstdefinestyle{mystyle}{
    backgroundcolor=\color{backcolour},   
    commentstyle=\color{codegreen},
    keywordstyle=\color{magenta},
    numberstyle=\tiny\color{codegray},
    stringstyle=\color{codepurple},
    basicstyle=\ttfamily\footnotesize,
    breakatwhitespace=false,         
    breaklines=true,                 
    captionpos=b,                    
    keepspaces=true,                 
    numbers=left,                    
    numbersep=5pt,                  
    showspaces=false,                
    showstringspaces=false,
    showtabs=false,                  
    tabsize=2
}

\copyrightyear{2025}
\acmYear{2025}
\setcopyright{acmcopyright}
\acmConference[HSCC '25]{28th ACM International Conference on Hybrid Systems: Computation and Control}{May 6--9 2025}{Irvine, California}
\acmBooktitle{28th ACM International Conference on Hybrid Systems: Computation and Control (HSCC '25), May 6--9 2025, Irvine, California}

\begin{document}

\title{SAVER: A Toolbox for Sampling-Based, Probabilistic Verification of Neural Networks}
\author{Vignesh Sivaramakrishnan}
\email{vigsiv@unm.edu}
\orcid{0000-0003-1127-9628}
\affiliation{%
  \institution{University of New Mexico}
  \city{Albuquerque}
  \state{New Mexico}
  \country{USA}
}
\author{Krishna C. Kalagarla}
\email{kalagarl@unm.edu}
\orcid{0000-0003-0618-8342}
\affiliation{%
  \institution{University of New Mexico}
  \city{Albuquerque}
  \state{New Mexico}
  \country{USA}
}
\author{Rosalyn Devonport}
\email{devonport@unm.edu}
\orcid{0000-0003-1913-5637}
\affiliation{%
  \institution{University of New Mexico}
  \city{Albuquerque}
  \state{New Mexico}
  \country{USA}
}
\author{Joshua Pilipovsky}
\email{jpilipovsky3@gatech.edu}
\orcid{0000-0001-6217-6042}
\affiliation{%
  \institution{Georgia Institute of Technology}
  \city{Atlanta}
  \state{Georgia}
  \country{USA}
}
\author{Panagiotis Tsiotras}
\email{tsiotras@gatech.edu}
\orcid{0000-0001-7563-4129}
\affiliation{%
  \institution{Georgia Institute of Technology}
  \city{Atlanta}
  \state{Georgia}
  \country{USA}
}
\author{Meeko Oishi}
\email{oishi@unm.edu}
\orcid{0000-0003-3722-8837}
\affiliation{%
  \institution{University of New Mexico}
  \city{Albuquerque}
  \state{New Mexico}
  \country{USA}
}

\begin{abstract}
    \noindent
    We present a neural network verification toolbox to 1) assess the probability of satisfaction of a constraint, and 2) synthesize a set expansion factor to achieve the probability of satisfaction. 
    Specifically, the tool box establishes with a user-specified level of confidence whether the output of the neural network for a given input distribution is likely to be contained within a given set. Should the tool determine that the given set cannot satisfy the likelihood constraint, the tool also implements an approach outlined in this paper to alter the constraint set to ensure that the user-defined satisfaction probability is achieved. The toolbox is comprised of sampling-based approaches which exploit the properties of signed distance function to define set containment. 
\end{abstract}


\maketitle
\section{Introduction}

Neural networks (NN) have shown great promise in myriad domains, often demonstrating performance on part with, or even surpassing, traditional methods.  They have even been embedded as elements of safety-critical systems, such as in quadrotor motion planning \cite{neural_lander,kaufmann2023champion}, autonomous driving \cite{autonomous_driving}, aircraft collision avoidance \cite{ACASX}, and aircraft taxiing \cite{taxinet}. 
However, the deployment of NNs in safety-critical domains introduces significant challenges, particularly regarding their robustness to input uncertainties and adversarial attacks. These challenges currently represent one of the primary barriers to the broader adoption of NNs in such domains. For instance, in contexts such as malware detection and sparse regression, existing works have shown that adversarial inputs can be crafted with relative ease to exploit vulnerabilities in NNs \cite{malware_detection_attacks,sparse_regression_attacks}.

\begin{figure}[t!]
    \centering
    \includegraphics[width=\linewidth]{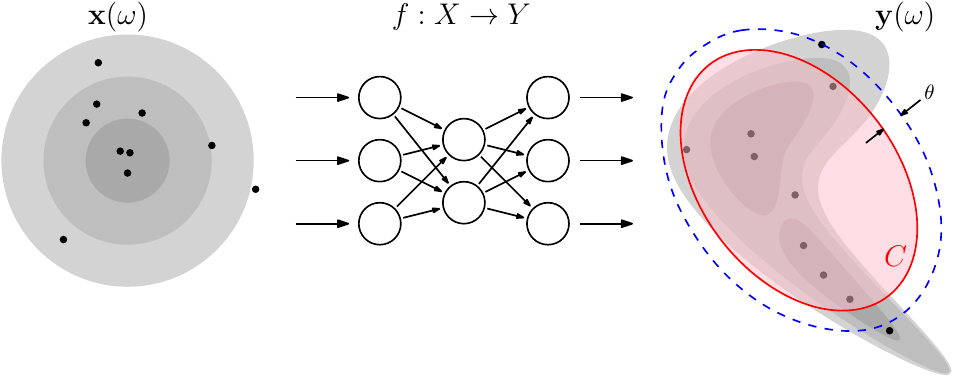}
    \caption{
    We consider neural nets with probabilistic inputs, and seek to identify the likelihood that the output distribution falls within a set $C$ with at least some probability $1-\Delta$ (Problem~\ref{prob:NNProbVerification}), and if necessary, to identify the smallest enlargement of the set $C$, by a distance $\theta$, in order to satisfy the same probability (Problem~\ref{prob:setEnlargment}).}
    \label{fig:verification-explainer}
\end{figure}

To address these concerns, the formal verification of NNs has emerged as a critical area of research. Broadly speaking, verification approaches can be categorized into two types: \textit{deterministic} and \textit{probabilistic}. Deterministic verification provides binary guarantees that the outputs of a NN satisfy a specific condition for all inputs within a defined set. This type of verification is crucial for worst-case safety guarantees. Several state-of-the-art techniques have been developed in this domain, including mixed-integer linear programming \cite{MILP1,MILP2,MILP3,MILP4}, satisfiability modulo theories \cite{SMT1,SMT2}, semidefinite programming \cite{SDP_exact1,SDP_exact2,SDP_exact3,SDP_exact4}, and reachability analysis \cite{reachability}. These methods typically assume a bounded input uncertainty set, centered around some nominal input.

On the other hand, probabilistic verification focuses on assessing the likelihood that a NN satisfies a safety condition under uncertain or random inputs, which may be unbounded. Such uncertainties can arise naturally from environmental noise, signal processing errors, or other exogenous disturbances. Furthermore, adversarial noise, even when imperceptible, can significantly alter NN outputs, particularly in deep networks for tasks like image classification \cite{random_attacks1,random_attacks2}.

Probabilistic verification addresses the key question: Given a random input vector $\mathbf{x}_0$ and a neural network $f$, what is the probability that the output $\mathbf{y} = f(\mathbf{x})$ lies within a predefined safety set $C$?
Specifically, we seek to ensure that this probability exceeds a certain user-defined threshold, expressed as
\begin{equation}~\label{eq:verif_certificate}
    \mathbb{P}_{\mathbf{y}}(C) \geq 1 - \Delta,
\end{equation}
where $\Delta \in (0,1)$ represents the satisfaction probability.

A major challenge in probabilistic NN verification is the difficulty of propagating input uncertainty through highly nonlinear mappings. Existing approaches to computing or approximating the chance constraint in (1) can be categorized as either \textit{analytical} or \textit{sampling-based}.
Analytical methods include the propagation of confidence ellipsoids using approximate networks, $\tilde{f}$, which reduces the verification problem to a semidefinite program \cite{PV1}, and the use of characteristic functions to propagate input distributions exactly in ReLU networks, drawing on concepts from discrete-time dynamical systems and Fourier transforms \cite{PV2}. The PROVEN framework \cite{PV3} approximates the safety probability using linear approximations and concentration inequalities for (sub-)Gaussian inputs with bounded support. Using similar ideas, other methods such as CC-Cert \cite{PV4} combine Cramer-Chernoff bounds with sample propagation to estimate the safety probability, while more recent approaches use branch-and-bound techniques to simultaneously upper and lower bound the probability of safety \cite{branch_and_bound}. Alternatively, scenario-based approaches compute safety certificates by solving chance-constrained programs based on the number of samples, either by directly lower-bounding the success probability \cite{scenario1} or by determining the largest norm ball $\mathbb{B}_{\varepsilon}$ that is likely to enclose the output \cite{scenario2}.

In this paper, we present SAVER: SAmpling-based VERification of Neural Nets, a Python toolbox for 
sampling-based probabilistic verification.
We use the  Dvoretzky-Kiefer-Wolfowitz Inequality~\cite{dvoretzky1956asymptotic} and scenario optimization approach~\cite{pmlr-v120-devonport20a}, which provide bounds on the number of samples necessary to validate a specification to a desired probability as in \eqref{eq:verif_certificate}.
To encode the set specifications, we define the sets as SDFs, whose properties allow us to expand the set should the original specification not achieve the satisfaction probability. 
We apply this approach on a three examples: 1) the containment of feedforward neural network's output, 2) the robustness of an image classifier, and 3) the resulting position of a aircraft with noise added to the neural network.


\section{Problem Formulation}

Let $(\Omega,\mathcal{M}(\Omega),P)$ be a probability tuple. 
The set $\Omega$ is the set of all possible outcomes, $\mathcal{M}(\Omega)$ is the set of events, i.e., a $\sigma$-algebra, where each event is a set of outcomes, and a function $P:\mathcal{M}(\Omega)\rightarrow [0,1]$ which assigns a probability to each set in the $\sigma$-algebra.
A random variable is a measurable function, $\mathbf{y}: \Omega \rightarrow \mathcal{Y}$.
We denote the space of random variables as $\mathbf{y}\in L^p(\Omega,\mathcal{M}(\Omega),P) = \left\{\mathbf{y}:\Omega\rightarrow \mathcal{Y}| \, \int_{\mathcal{Y}}|\mathbf{y}(\omega)|^p\ \mathrm{d}P(\omega)<\infty\right\}$.
The probability that $\mathbf{y}$ will take on a value in $S$ we represent by a probability measure, $P_{_\mathbf{y}}(S) = P(\mathbf{y}^{-1}(S))=P\left(\left\{ \omega\in\Omega | \mathbf{y}(\omega)\in S\right\}\right)$ for $S \in\mathcal{M}(\mathcal{Y})$.
We denote a constraint set $C\in\mathcal{M}(\mathcal{Y})$ through a measurable signed distance function (SDF) $g_{C}:\mathcal{Y}\rightarrow\mathbb{R}$,
\begin{equation}
\label{eq:signed_distance}
	g_C(y) \defeq 
	\begin{cases}
		\hspace{0.95em}\inf_{p\in C}\|p-y\|,\ y\in \mathcal{Y}\backslash C,\\
	   -\inf_{p\in \mathcal{Y}\backslash C}\|p-y\|,\ y\in C.\\
	\end{cases}
\end{equation}
Put simply, the SDF says that the point $y\in\mathcal{Y}$ is within the set if its output is negative or zero and positive otherwise. Set operations such as union, intersection, and set difference are readily captured using SDFs. We use SDFs to formulate the two NN verification problems the toolbox solves, stated formally here and depicted graphically in Figure~\ref{fig:verification-explainer}.

\begin{problem}
    \label{prob:NNProbVerification}
    Given a neural network, $f:\mathcal{X}\rightarrow\mathcal{Y}$, a SDF $g_C: \mathcal{Y} \rightarrow \mathbb{R}$, and a probability of satisfaction $\Delta\in (0,1)$, determine if
    \begin{equation}
        \label{eq:probOfSatisfaction}
        \mathbb{P}\left(\{\omega: g_C(f(\mathbf{x}(\omega)))\leq 0\}\right)\geq 1-\Delta,
    \end{equation}
    where $1-\Delta\in(0,1)$ is the probability of satisfaction.
\end{problem}

\begin{problem}
    \label{prob:setEnlargment}
    Given a neural network, $f:\mathcal{X}\rightarrow\mathcal{Y}$, a SDF $g_C: \mathcal{Y} \rightarrow \mathbb{R}$, and a probability of satisfaction $\Delta\in (0,1)$, solve the optimization problem,
    \begin{subequations}
    \label{opt:levelSet}
    \begin{align}
        \underset{\theta\in\mathbb{R}}{\mathrm{minimize}} &\quad \theta,\\
        \mathrm{subject\ to} &\quad \mathbb{P}\left(\{\omega: g_{C}(f(\mathbf{x}(\omega)))-\theta\leq 0\}\right)\geq 1-\Delta. \label{opt:levelSetb}
    \end{align}
    \end{subequations}
    That is, we find a new set $C^* = \{y\in\mathcal{Y}: g_C(y) - \theta^* \leq 0\} $, where $\theta^*$ is the optimal solution of \eqref{opt:levelSet}, such that it satisfies \eqref{prob:NNProbVerification}.
\end{problem}

\begin{rem}
    Note that both Problem~\ref{prob:NNProbVerification} and \ref{prob:setEnlargment} are focused on probabilistic verify neural networks. 
    However, this formulation is generic, and so $f$ could also represent a dynamical system with a neural network in the loop, or other functions.
\end{rem}

Solving Problem~\ref{prob:NNProbVerification} is crucial for determining whether neural network or a system with a neural network in the loop satisfies specifications with a given probability. 
We use Problem~\ref{prob:setEnlargment} for:
\begin{enumerate}
    \item enlargement: identify how much larger set must be to ensure the satisfaction likelihood $1-\Delta$
    \item reduction: shrink the set $C$, in the event that the desired specification can be satisfied with high probability.
\end{enumerate}

\section{Sampling-Based Verification}

To evaluate the expression on the left hand side of the inequality in \eqref{eq:probOfSatisfaction}, we rewrite the probability of neural network output residing in the set as a cumulative distribution function (CDF), 
\begin{equation} \label{eq:ogProbConstraint}
    \begin{split}
        \Phi_{g_C(f(\mathbf{x}))}(y) &= 
        \mathbb{E}[\mathbf{1}_{\{g_C(f(\mathbf{x}))\leq y\}}], \\ 
        &= \mathbb{P}(\{\omega\in\Omega|\ g_C(f(\mathbf{x}(\omega))) \leq y\}).
    \end{split}
\end{equation}
We can approximate \eqref{eq:ogProbConstraint} 
empirically as, 
\begin{equation}
    \hat\Phi_{g_C(f(\mathbf{x}))}(y)= 
        \frac{1}{N}\sum_{i=1}^N\mathbf{1}_{\{g_C(f(\mathbf{x}^{(i)}))\leq y\}}.\label{eq:empiricalCDF}
\end{equation}

\subsection{Dvoretzky-Kiefer-Wolfowitz Inequality}\label{sec:DKW}

One way to determine how many samples are needed to estimate \eqref{eq:ogProbConstraint} via \eqref{eq:empiricalCDF} is via the Dvoretzky-Kiefer-Wolfowitz (DKW) inequality~\cite{dvoretzky1956asymptotic}, which provides a bound on the difference between the empirical distribution function  $\hat\Phi_{g_C(f(\mathbf{x}))}$ and the true CDF $\Phi_{g_C(f(\mathbf{x}))}$. Formally, given $N$ i.i.d. samples, the DKW inequality states that for any error tolerance $\epsilon > 0$,
\begin{equation}\label{eq:sampleReqDKW}
\mathbb{P}\left(\sup_x |\hat\Phi_{g_C(f(\mathbf{x}))}(x) - \Phi_{g_C(f(\mathbf{x}))}(x)| > \epsilon\right) \leq 2 \mathrm{e}^{-2N\epsilon^2}.
\end{equation}
Further, to determine the number of samples $N$ an obtain an empirical CDF from \eqref{eq:empiricalCDF}
that is within $\epsilon$ to the actual one in \eqref{eq:ogProbConstraint} with confidence level $1-\beta$, we can rearrange \eqref{eq:sampleReqDKW} to obtain, 
\begin{equation}
\label{eq:sampleBoundDKW}
N \geq \left\lceil -\frac{1}{2\epsilon^2}\ln\left(\frac{\beta}{2}\right) \right\rceil.
\end{equation}

Therefore, solving Problem~\ref{prob:NNProbVerification} is merely a matter of using \eqref{eq:sampleReqDKW} to compute the number of samples via a user specified $\epsilon,\ \beta\in(0,1)$. 
For Problem~\ref{prob:setEnlargment}, we first note that \eqref{opt:levelSet} is equivalent to the definition of the quantile function. 
\begin{defn}
The quantile function $Q: [0,1] \rightarrow \mathbb{R}$ is the inverse of the CDF, provided that $\Phi_{g_C(f(\mathbf{x}))}(x)$ is continuous and non-decreasing. 
Formally, for $p \in [0, 1]$, the quantile function is given by
\begin{equation}
\label{eq:quantileCDF}
Q(p) = \inf \{ x \in \mathbb{R} : \Phi_{g_C(f(\mathbf{x}))}(x) \geq p \}.
\end{equation}
In other words, $Q(p)$ returns the value $x$ such that the probability of a random variable being less than or equal to $x$ is at least $p$.
\end{defn}

We can even use root-finding, such as the bisection algorithm, to find the smallest $\theta$ such that, 
\begin{equation}
\label{eq:dkwEnlargment}
    \hat\Phi_{g_C(f(\mathbf{x}))}(\theta^*) = 1-\Delta.
\end{equation}

\subsection{Scenario Optimization}\label{sec:scenario}

Scenario optimization constructs data-driven approximations of solutions for the chance-constrained optimization problem
\begin{subequations}
\label{eq:ccScenarioOG}
    \begin{align}
        \underset{\theta\in\mathbb{R}^{n_{\theta}}}{\mathrm{minimize}} &\quad c^\intercal \theta\\
        \mathrm{subject\ to} &\quad \mathbb{P}\left(\{\omega: h(\mathbf{x}(\omega),\theta)\leq 0\}\right)\geq 1-\Delta,
    \end{align}
\end{subequations}
where $h:\mathbb{R}^n\times \mathbb{R}^{n_{\theta}}\rightarrow\mathbb{R}$ is convex with respect to its second argument $\theta$.
Instead of directly evaluating the chance constraint, which is generally impractical,
the scenario approach approximately enforces the constraint in \eqref{eq:ccScenarioOG} by utilizing samples of $\mathbf{x}$, yielding the scenario relaxation
\begin{subequations}\label{opt:scenarioProblem}
    \begin{align}
        \underset{\theta\in\mathbb{R}^{n_{\theta}}}{\mathrm{minimize}} &\quad c^\intercal \theta\\
        \mathrm{subject\ to} &\quad h\left(\mathbf{x}^{(i)},\theta\right)\leq 0,\quad i=1,\hdots,N. \label{eq:scenarioCC}
    \end{align}
\end{subequations}
An optimal solution $\theta^*$ of~\eqref{opt:scenarioProblem} will not be an optimizer of~\eqref{eq:ccScenarioOG} due to the sampling approximation since the set of optimizers is generally of measure zero; however, the set of feasible solutions, i.e., values of $\theta$ that satisfy the chance constraint while not necessarily minimizing the objective, has positive measure that we can bound. 
In particular, a sufficient condition that an optimal solution of~\eqref{opt:scenarioProblem} be a feasible solution of~\eqref{eq:ccScenarioOG} with probability $\ge 1-\beta$ is that the number of samples $N$ satisfies the bound
\begin{equation}\label{eq:scenarioSamples}
    N \geq \left\lceil\frac{1}{\Delta} \left( \frac{\mathrm{e}}{\mathrm{e} - 1} \right) \left( \log \frac{1}{\beta} + n_{\theta} \right)\right\rceil,
\end{equation}
where $\beta\in(0,1)$ is the confidence parameter, and $n_{\theta}$ is the number of decision variables. 
With a user-specified confidence parameter $\beta$ and the number of decision variables
$n_\theta$, we have that the optimal solution $\theta^*$ results in satisfaction of the chance constraint with probability $1-\Delta$ with confidence $1-\beta$~\cite{pmlr-v120-devonport20a}. 

For the neural network verification problem, we can pose the optimization problem in \eqref{opt:levelSet} as follows,
    \begin{subequations}
    \label{opt:levelSetScenario}
    \begin{align}
        \underset{\theta\in\mathbb{R}}{\mathrm{minimize}} &\quad \theta,\\
        \mathrm{subject\ to} &\quad g_{C}\left(f\left(\mathbf{x}^{(i)}\right)\right)-\theta\leq 0,\quad i=1,\hdots,N,
    \end{align}
    \end{subequations}
where we can determine the desired number of samples by setting $n_{\theta}$ to 1.
Note that, if $\theta^* \leq 0$, then we satisfy the specification, thereby solving Problem~\ref{prob:NNProbVerification}. 
Should $\theta^* > 0$, then we address Problem~\ref{prob:setEnlargment}, where we determine the expansion of the set to ensure $1-\Delta$ constraint satisfaction.  

Note that the optimization problem in \eqref{opt:levelSetScenario} is convex, because the decision variable $\theta$ is linear. 
Further, the optimization problem does not require the convexity of the set specified by the SDF. 
We can solve the optimization problem by Algorithm 1.

\begin{algorithm}[ht!]
\caption{Scenario-Based Verification Algorithm}
\label{alg:sample}
\begin{algorithmic}[1]
\REQUIRE Confidence parameter, $1\beta$; probability of satisfaction, $\Delta$; input distribution, $\mathbb{P}_\mathbf{x}$; nonlinearity, such as neural network or system, $f:\mathbb{R}^n\rightarrow \mathbb{R}^m$;
specification, $g_C: \mathbb{R}^m\rightarrow\mathbb{R}$
\ENSURE Set modification, $\theta^*$
\STATE Determine the number samples, $N$, via \eqref{eq:scenarioSamples}. 
\STATE Sample $N$ samples of $\mathbf{x}$, i.e. $x^{(i)}\sim\mathbb{P}_\mathbf{x}$ for $i=1,\hdots,N$.
\STATE Compute $y^{(i)} = g_C\left(f\left(x^{(i)}\right)\right)$ for $i=1,\hdots,N$.
\STATE Find the largest $y^{(i)}$ and set $\theta^* = y^{(i)}$. 
\RETURN $\theta^*$
\end{algorithmic}
\end{algorithm}


\section{Toolbox Features}
\begin{figure}[ht!]
    \centering

\begin{tikzpicture}[x=1in, y=1in]
    \tikzstyle{block} = [rectangle, draw, minimum width=1mm, minimum height=1mm,text width=1.6cm,align=center]
    \tikzstyle{longBlock} = [rectangle, draw, minimum width=1mm, minimum height=1mm,text width=3cm,align=center]
    \tikzstyle{arrow} = [->, thick]

    \node[block] (init) at (0.15, 0.5) {Initialize Verification\\ {\scriptsize Section~\ref{sec:initVerif}}};
    \node[longBlock] (verif) at (1.5, 0.5) {\texttt{verficationProblem}};
    \node[block] (specification) at (1.5, 0) {Specification\\ {\scriptsize Section~\ref{sec:SDF}}};
    \node[block] (sampler) at (1.5, 1.145) {Sample Assessment\\ {\scriptsize Section~\ref{sec:samplerVerif}}};
    \node[] (sReq) at (0.95,0.71) {{\scriptsize \texttt{samplesRequired()}}};
    \node[] (samples) at (1.9,0.71) {{\scriptsize \texttt{samples()}}};
    \node[] (specAdd) at (1.85,0.3) {{\scriptsize \texttt{specification()}}};

    \node[] (evalFunc) at (2.7,0.71) {{\scriptsize \texttt{probability()}}};
    \node[] (samples) at (2.7,0.3) {{\scriptsize \texttt{modifySet()}}};

    \node[block] (eval) at (2.75,1.12) {Likelihood Evaluation\\ {\scriptsize Section~\ref{sec:satisfiedProb}}};
    \node[block] (optim) at (2.75,-0.05) {Set enlargement\\ {\scriptsize Section~\ref{sec:modifySet}}};
    
    \draw[-{Latex}] (init) -- (verif);
    \draw[-{Latex}] (specification) -- (verif);
    \draw[-{Latex}] (verif.east) -- (eval);
    \draw[-{Latex}] (verif.east) -- (optim);
    \draw[-{Latex}] (1.35,0.58) -- ++(0,0.32);
    \draw[-{Latex}] (1.65,0.9) -- ++(0,-0.32);
    
\end{tikzpicture}
    \caption{Flowchart depicting SAVER toolbox elements.}
    \label{fig:toolboxFlow}
\end{figure}
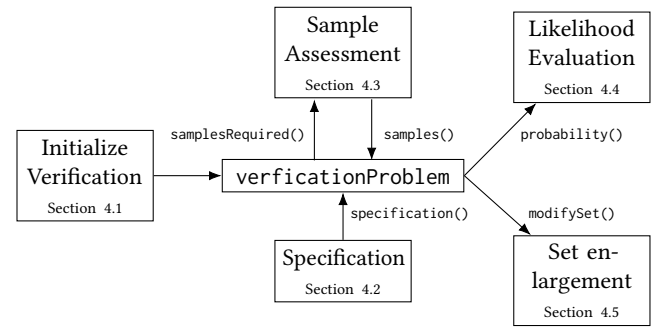
The toolbox is downloadable from \url{https://bit.ly/SAVERtoolbox}.  
It runs on Python and requires the following packages: \texttt{torch}, \texttt{torchvision}, \texttt{numpy}, \texttt{matplotlib}, \texttt{cvxpy}, \texttt{scipy} and \texttt{jupyterlab}. 
We implement the toolbox in Python as it contains a number of popular packages for neural network training such as, but not limited to, TensorFlow~\cite{tensorflow2015-whitepaper}, PyTorch~\cite{NEURIPS2019_9015}, and JAX~\cite{jax2018github}.
By default, mathematical operations are handled by NumPy~\cite{harris2020array}. 
Figure~\ref{fig:toolboxFlow} overviews the functions within our approach and how one would go about setting up a verification task.

\subsection{Initializing a Verification Problem}\label{sec:initVerif}
We can start a sampling-based verification problem by the following functions:
\begin{lstlisting}[language=Python]
# Setup code omitted
verifDKW = verify.usingDKW(betaDKW,epsilonDKW,Delta)
verifScenario = verify.usingScenario(betaScenario,Delta)
\end{lstlisting}

Here, \texttt{verify.usingDKW} and \texttt{verify.usingScenario} initialize sampling-based verification problem using DKW or scenario optimization respectively.
Both functions take \texttt{Delta} corresponding to $\Delta$, the user-defined satisfaction probability.
In the function \texttt{verify.usingDKW}, the input variables \texttt{betaDKW} and \texttt{epsilonDKW}, and \texttt{Delta} correspond to $\beta$ and $\epsilon$ respectively which appear in \eqref{eq:sampleBoundDKW}. 
In function \texttt{verify.usingScenario}, the input variables \texttt{betaScenario} corresponds to $\beta$ in Section~\ref{sec:scenario}, the confidence for which the probability of satisfaction is greater than $1-\Delta$.

\subsection{Adding a Specification}\label{sec:SDF}
\begin{figure}[ht!]
    \centering
    \includegraphics[width=\linewidth]{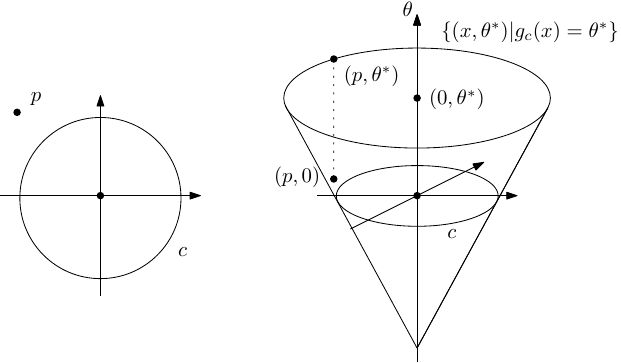}
    \caption{A subset set $C$ (left) of a vector space $\mathcal{X}$ and the graph of its SDF $g_C$ (right). A point $p$ positioned at a signed distance $\theta^*$ away from $C$ lies on the border of the enlarged set $C^* = \{x:\ g_C(x) = \theta^*\}$, which is the projection of the graph section $\{(x,\theta^*):(g_C(x) = \theta^*)\}$ onto $\mathcal{X}$. Here $\theta^* > 0$, resulting in an enlargement; if $\theta^*$ were negative, the new set would be a reduction.}
    \label{fig:twoNormBall}
    \Description[<short description>]{<long description>}
\end{figure}

The toolbox consists of two SDFs: 1) norm-ball and 2) polytopes.
We present the norm-ball with an example implementation of a two norm-ball (Figure~\ref{fig:twoNormBall}),
\begin{equation}\label{eq:twoNormBall}
    g_{C}(p) = ||p - c||_2 - r.
\end{equation}
The function which implements this in our toolbox is:
\begin{lstlisting}[language=Python]
# Define SDF:
normSDF = sdfs.norm(center,zeroRadius,norm=2)
\end{lstlisting}
The function \texttt{sdfs.norm} requires variables \texttt{center} and \texttt{zeroRadius} which correspond to $c$ and $r$ in \eqref{eq:twoNormBall}.
The function utilizes "\texttt{norm=}" variable to define what norm the SDF is, and is a two norm by default. 
For sake of space, we do not expand on the implementation of the polytope SDF except in subsection~\ref{sec:imageClassifer} where we utilize it for image robustness.
A SDF is added to the verification task calling the following functions with \texttt{normSDF} as an input argument: 
\begin{lstlisting}[language=Python]
# Add SDF: 
verifDKW.addSpecification(normSDF)
verifScenario.addSpecification(normSDF)
\end{lstlisting}
Should the user desire to change the specification in the verification problem, calling \texttt{addSpecification} from the verification problem with a new input variable for the specification, e.g., \texttt{newSDF}.

\subsection{Adding Samples}\label{sec:samplerVerif}
Note that for this toolbox, the samples must come from a user-defined sampling function or simulator. 
To provide the number of samples to the sampler or simulator, we provide the user with the following functions which output the number of samples: 
\begin{lstlisting}[language=Python]
# How many samples are needed: 
verifDKW.samplesRequired()
verifScenario.samplesRequired()
\end{lstlisting}
We add samples to verification problem by evaluate the following function with the variable \texttt{outputSamples}: 
\begin{lstlisting}[language=Python]
# Add samples to the verification problem: 
verifDKW.samples(outputSamples)
verifScenario.samples(outputSamples)
\end{lstlisting}
Should the user desire to change samples for the verification problem, calling \texttt{samples} from the verification problem with a new input variable of samples, e.g. \texttt{newOutputSamples}.

\subsection{Check Probabilistic Satisfaction of Specification}\label{sec:satisfiedProb}
To address Problem~\ref{prob:NNProbVerification}, we can call the following function for either sampling-based approach: 
\begin{lstlisting}[language=Python]
# Check if the samples satisfy the specification: 
verifDKW.probability()
verifScenario.probability()
\end{lstlisting}

\subsection{Modify the Specification to Ensure Probabilistic Satisfaction}\label{sec:modifySet}
Should we not be able probabilistically satisfy the specification, we can solve Problem~\ref{prob:setEnlargment} to determine what
how much the set must grow to ensure the satisfaction probability. 
\begin{lstlisting}[language=Python]
# Modify the specification:
verifDKW.modifySet()
verifScenario.modifySet()
\end{lstlisting}


\section{Examples}

We present three examples that highlight the usage of our sampling-based verification approach.
All experiments were run on a Intel 10900K with 128GB RAM running PopOS 22.04 with Python 3.10.12.
The examples in Sections~\ref{sec:FeedforwardNN} and~\ref{sec:imageClassifer} utilize PyTorch to conduct evaluations of the neural network and are then converted to NumPy arrays.

\subsection{Feedforward Neural Network with ReLU Activations}\label{sec:FeedforwardNN}
\begin{figure}[ht!]
    \centering
\begin{tikzpicture}[x=2.2cm,y=2.2cm]
\node[] at (-0.5,2.75) {\includegraphics[width=1in]{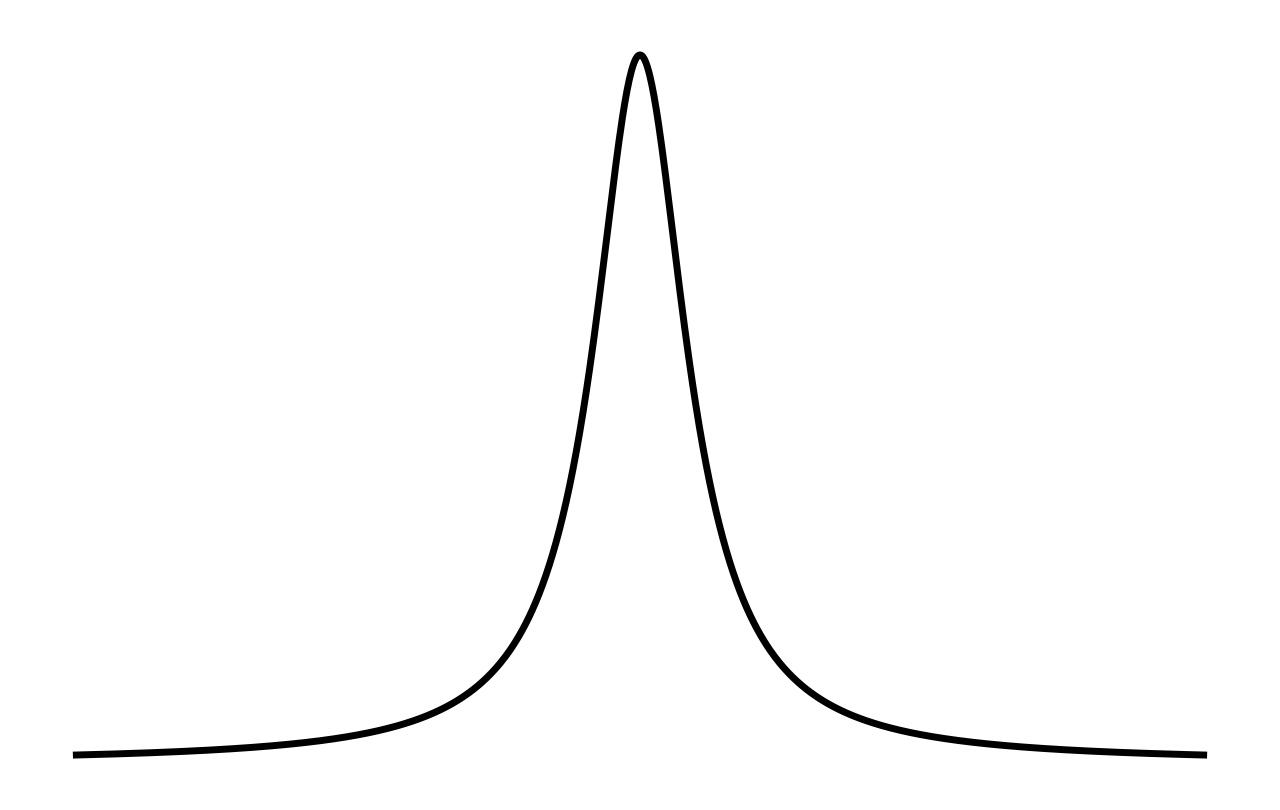}};
\node[] at (-0.5,2) {$\mathbf{x}\in \mathbb{R}^4$};
\node[] at (-0.5,1.8) {where $\mathbf{x}_i\sim \mathrm{Cauchy}(0,1)$};
\node[] at (1.65,2.75) {\includegraphics[width=1in]{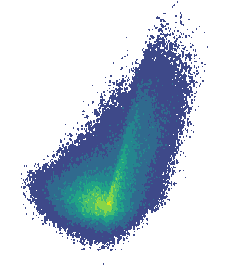}};
\node[] at (1.65,2) {$\mathbb{P}_{\mathbf{y}}(C) \geq 1 - \Delta$ ?};
\node[] at (1.65,1.8) {$\mathbf{y}\in \mathbb{R}^2$};
\draw[red,thick] (1.65,2.5) circle [radius=0.3];
\begin{scope}[scale=0.25]
  \readlist\Nnod{4,10,10,2} 
  \foreachitem \N \in \Nnod{ 
    \foreach \i [evaluate={\x=\Ncnt; \y=\N/2-\i+0.5; \prev=int(\Ncnt-1);}] in {1,...,\N}{ 
      \node[mynode] (N\Ncnt-\i) at (\x,0.65*\y+10) {};
      \ifnum\Ncnt>1 
        \foreach \j in {1,...,\Nnod[\prev]}{ 
          \draw[thin,draw=black!40] (N\prev-\j) -- (N\Ncnt-\i); 
        }
      \fi 
    }
  }
  \end{scope}
\end{tikzpicture}
    \caption{We propagate a standard Cauchy distribution at the input through a feedforward neural network with ReLU activation functions and determine whether the output resides within the set $C$ (in red) with probability greater than $1-\Delta$.}
    \label{fig:FeedforwardNNVisual}
    \Description[<short description>]{<long description>}
\end{figure}
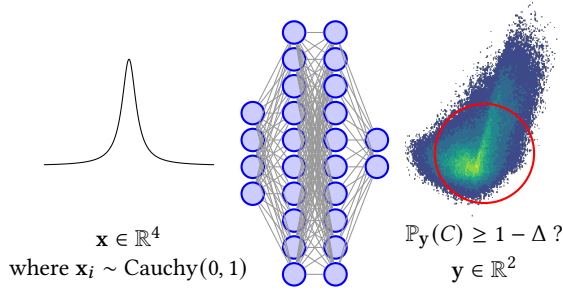

We employ our sampling-based verification tool on a simple feedforward neural network with ReLU activation functions similar to ~\cite{PV2}.
As we show in Figure~\ref{fig:FeedforwardNNVisual}, the network takes in samples from a standardized Cauchy distribution, a distribution with undefined moments, passes through the network which consists 4 inputs, 2 layers, each with 10 neurons, and 2 outputs. 
The set specification we wish to ensure for a satisfaction probability of $1-\Delta = 0.999$ is a two norm ball of radius 20,000, 
\begin{equation}\label{eq:ffNNSetSpecification}
    C = \{y\in\mathbb{R}^{2}: ||y||_2 \leq 20,000\}.
\end{equation}
This results in a SDF, 
\begin{equation}
    g_{C}(p) = \|p\|_2 - 20,000.
\end{equation}
Since a Cauchy distribution is heavy tailed distribution, we set a conservative estimate of how large the set must be, in hopes of reducing the set.

\begin{figure}[ht!]
    \centering
    \includegraphics[width=\linewidth]{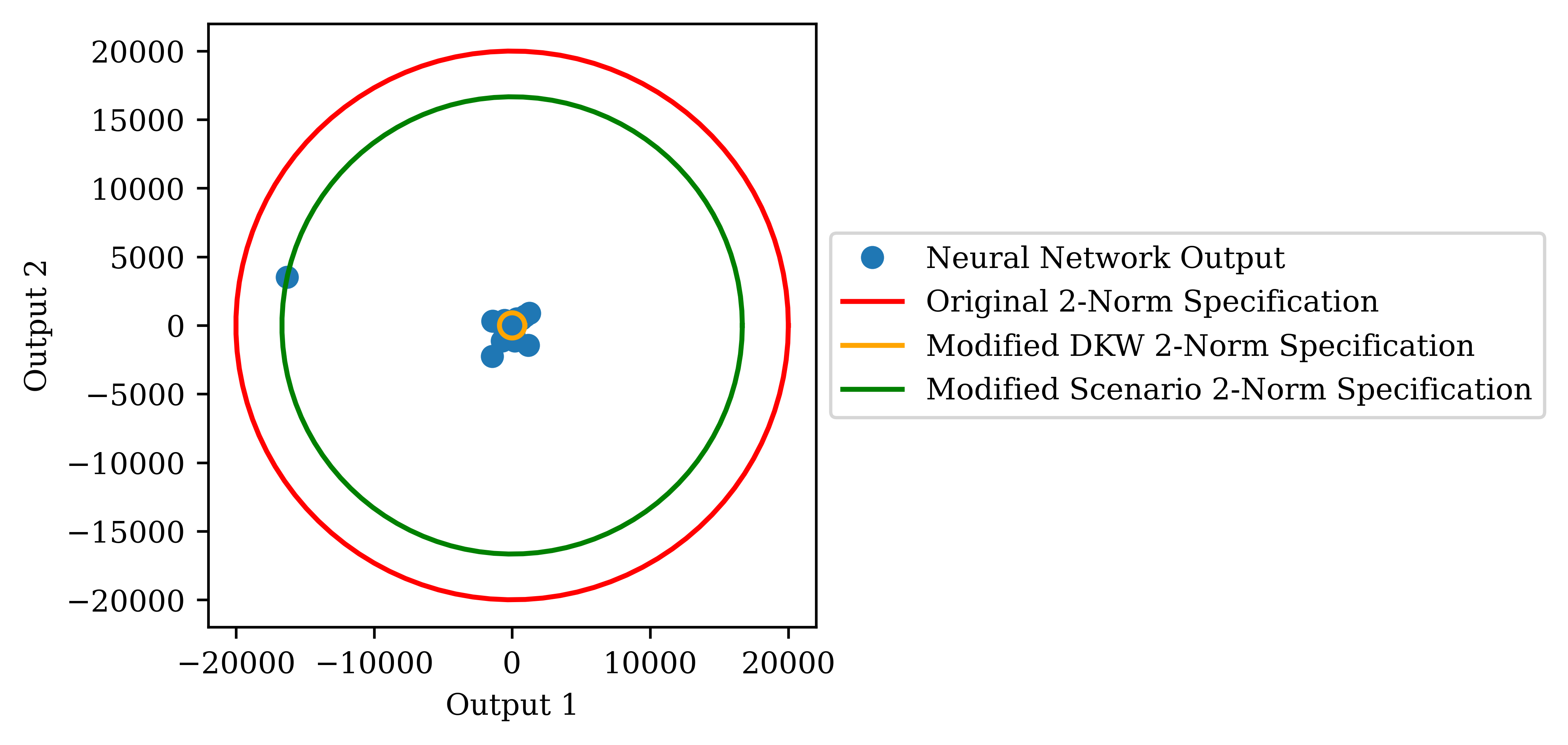}
    \caption{Samples of the output of the neural network are superimposed on the set specification \eqref{eq:ffNNSetSpecification}.
    Through both DKW and scenario optimization approaches, we achieve probabilistic satisfaction of the specification with probability greater than 99\%. 
    To reduce the set's conservatism, we revise the specification using DKW (yellow) and scenario (green) approaches to solve Problem~\ref{prob:setEnlargment} while ensuring 99.9\% specification satisfaction.}
    \label{fig:FeedforwardNNSamples}
    \Description[<short description>]{<long description>}
\end{figure}

We run both the DKW and scenario-based approaches from our toolbox and present the results in Figure~\ref{fig:FeedforwardNNSamples}, along with the samples. 
For the DKW approach, we utilize a confidence, $1-\beta$, of 0.999 and an error from the true CDF, $\epsilon$, of 0.001, resulting in requiring 3800452 samples.
The scenario approach uses the same confidence as DKW, requiring $12,510$ samples for the verification task.
The analysis we conduct with both the DKW and scenario approaches indicate that we do satisfy the specification with 0.999 probability, thereby addressing Problem~\ref{prob:NNProbVerification}. 
Nonetheless, to reduce our initial estimate of the set,  we solve \eqref{opt:levelSet} in Problem~\ref{prob:setEnlargment} with both the DKW and scenario approach which result in $\theta^*_{\rm DKW} = -19232.62$ and $\theta^*_{\rm Scenario} = -3779.28$ respectively.
Specifically, the DKW-based approach results in a reduction relative to an empirical CDF with known error relative to the true CDF that holds with a specific confidence. 
In contrast, the scenario-based approach results in a larger set expansion as it prioritizes probabilistic satisfaction of the specification with user-defined confidence. 

\subsection{Image Classifier}\label{sec:imageClassifer}

\begin{figure}[ht!]
    \centering
    \includegraphics[width=\linewidth]{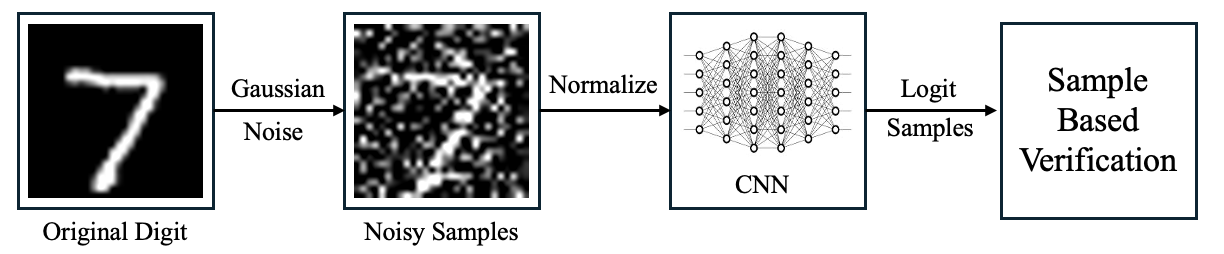}
    \caption{We determine the robustness of our classifier by perturbing each pixel of a fixed digit from the MNIST database with standard Gaussian with a noise variance of 0.25, i.e. $\mathcal{N}(0,0.25)$.} 
    \label{fig:cnnExperiment}
    \Description[<short description>]{<long description>}
\end{figure}

We also validated the sampling-based approach by examining the robustness of a  Convolutional Neural Network (CNN) classifier trained on a normalized MNIST dataset ~\cite{deng2012mnist}. Specifically, we evaluated whether the classifier maintains a high probability of correct classification on noisy versions of an originally correctly classified image of the digit $7$.

To formalize the requirement of correct classification, we define a constraint set $C$ in the logit space $y \in \mathbb{R}^{10} $ of the classifier. Logits are the outputs from the last layer of a neural network before applying the softmax function, representing the raw, unnormalized class scores. The constraint set is a polytope in the logit space which specifies that the classifier's output is `7'. Let $ y = f(x) $ denote the logit vector output by the CNN for a normalized input image $ x $, where $y_i$ is the logit corresponding to class $i$. The condition for the correct classification as `7' is ensured by enforcing that the logit for class `7' (denoted as $y_7$) is the highest among all classes. In matrix form, these constraints can be expressed as:
\begin{equation}\label{eq:safe_cnn}
    C = \{y \in \mathbb{R}^{10}: A y \leq b\}.
\end{equation}
where each row in \( A \) includes a \( +1 \) for \( y_7 \), a \( -1 \) for a particular \( y_i \) (where \( i \neq 7 \)), and \( 0 \) elsewhere, \( b \) is a zero vector. 

 \begin{figure*}[t!]
    \centering
    \begin{tikzpicture}[auto, node distance=2cm]
        \node[draw, rectangle] (image) {\includegraphics[width=2cm]{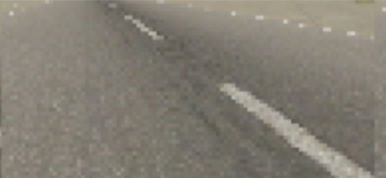}};
        \node[above of=image, node distance=1cm]{Image};
        
        \node[draw, rectangle, right of=image, node distance=3.5cm] (downsampled) {\includegraphics[width=2cm]{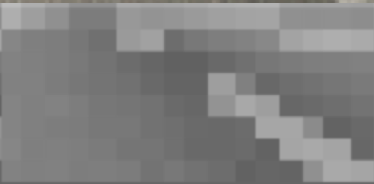}};
        \node[below of=downsampled, node distance=1cm]{Downsampled Image};
        
        \node[draw, rectangle, right of=downsampled, node distance=3.5cm] (nn) {TaxiNet};
        \node[draw, rectangle, right of=nn, node distance=3.5cm] (xplane) {XPlane Simulator};
        
        \draw[-{Latex}] (image) -- (downsampled);
        \draw[-{Latex}] (downsampled) -- (nn);
        \draw[-{Latex}] (nn) -- (xplane);
        \draw[-{Latex}] (xplane) |- ++(0,-1.3) -| (image);
        
        \draw[-{Latex}] (downsampled) ++(0,1) node[above] {Noise, $\mathbf{w}\sim\mathcal{N}(0,50)$} -- (downsampled);
        
    \end{tikzpicture}
    \caption{The TaxiNet simulator represents a Cessna 208B Grand Caravan with a camera is placed under the right wing of the aircraft. 
   Gaussian noise is added as a disturbance to the downsampled camera image, and then the corrupted image is fed into a feedforward neural network designed to regulate cross track deviations, so that the aircraft tracks along the runway centerline while taxiing.
    }
    \label{fig:taxinetOverview}
    \Description[xxx]{xxx}
\end{figure*}
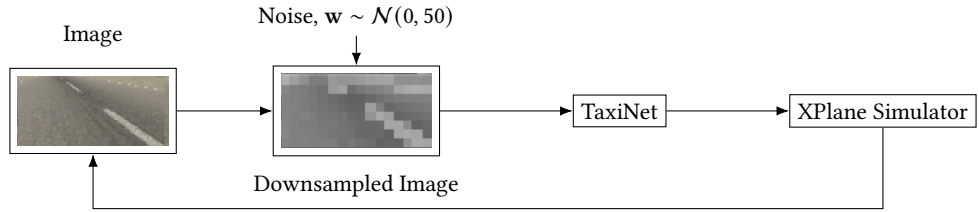

We then compute the SDF $g_C$ for the polytope $C$. This function quantifies how far the logit vector $y$ is from the boundary of $C$, providing a measure of classification stability:

\begin{equation}
    g_C(y) = 
\begin{cases} 
\min_{z \in C} \| y - z \|_2, & \text{if } y \notin C \\
-\min_{i} \frac{A_i y - b_i}{\| A_i \|_2}, & \text{if } y \in C 
\end{cases}
\end{equation}
 where \( A_i \) is the \(i\)-th row of the matrix \( A \). 

 \begin{itemize}
 \item If \( y \notin C \), the signed distance is positive and is computed as the shortest Euclidean distance to the polytope's boundary. This distance can be determined by solving the quadratic programming problem outlined above.
 \item If \( y \in C \), the signed distance is negative. It is determined by computing the perpendicular distances from \( y \) to each hyperplane defining the polytope's facets. The minimum of these distances represents the shortest distance to the boundary.
 \end{itemize}
 
We generated input samples by adding standard Gaussian noise with a mean of zero and a variance of $0.25$, thus distorting the digit (Figure~\ref{fig:cnnExperiment}). This pixel-wise noise is added to the un-normalized image, which is then normalized before CNN classification.
We require that the specification in \eqref{eq:safe_cnn} be satisfied with $1-\Delta=0.99$ probability. 
To determine if the specification is satisfied via DKW, we specify $\epsilon=0.01$ and $\beta=0.001$ for \eqref{eq:sampleReqDKW}, resulting in requiring $38,005$ samples. 
Likewise, we generated samples needed for the scenario approach with the same confidence as the DKW-based approach, requiring $1,251$ samples. 
We generated these samples by the above-described process. This resulted in determining that, by Problem~\ref{prob:NNProbVerification}, we
did not satisfy the specification in \eqref{eq:safe_cnn}. 

\subsection{TaxiNet}

We validated the sampling-based approach on the TaxiNet benchmark shown in Figure~\ref{fig:taxinetOverview}. 
TaxiNet consists of a neural network which predicts heading angle and the cross track position from images from a camera attached on the right wing of a Cessna 208B Grand Caravan taxiing at 5 m/s down runway 04 of Grant County International Airport~\cite{Katz2021}.
The crosstrack and heading angle are fed into a proportional controller,
\begin{equation}
\phi = -0.74p - 0.44\theta,
\end{equation}
where $\phi$ is the steering angle of the aircraft.
Gaussian noise to each pixel, i.e. $\mathbf{w}\sim\mathcal{N}(0,50)$, corrupts the downsampled image that is fed into TaxiNet.

\begin{figure}[ht!]
    \centering
    \includegraphics[width=\linewidth]{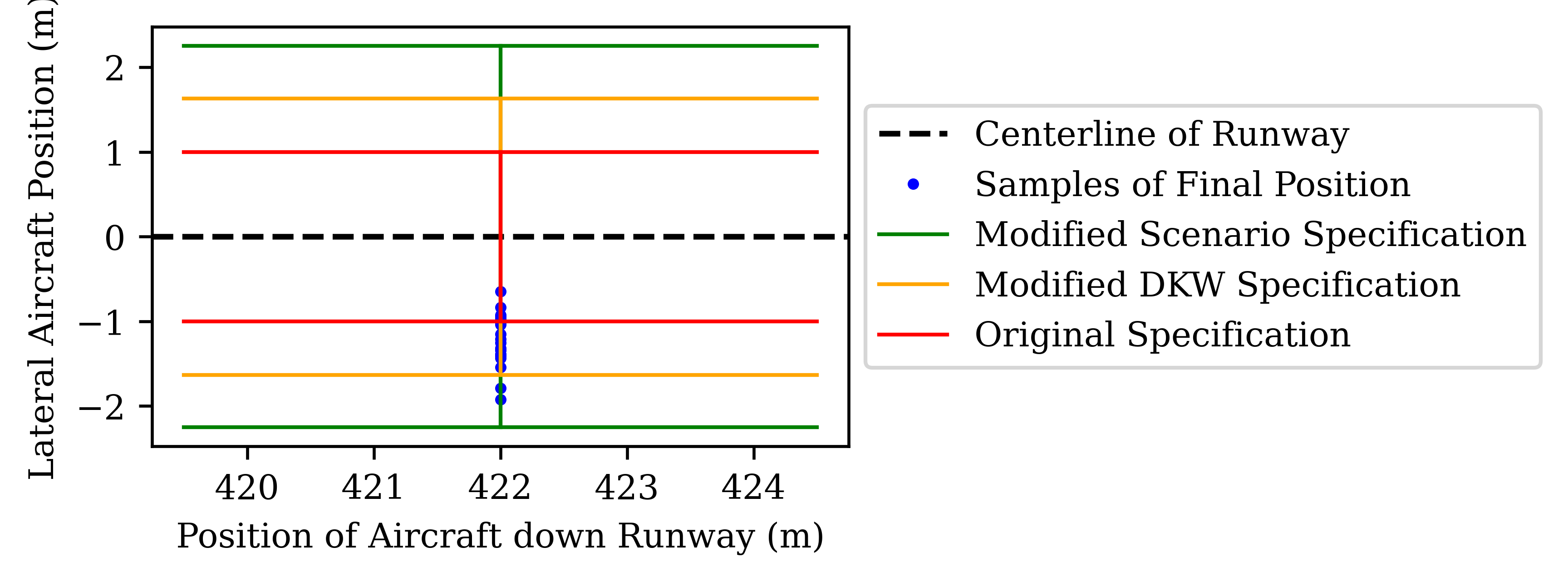}
    \Description[xxx]{xxx}
    \caption{Lateral view of the TaxiNet experiment. 
    The aircraft has an initial cross track deviation of 5 meters, and then taxis 100 meters along the runway. 
    However, solving Problem 1 reveals that because the downsampled images are corrupted, the aircraft cannot regulate its position to within 1 meter of the centerline with at least 90\% likelihood. 
    Solving Problem~\ref{prob:setEnlargment} results in enlargements when using either DKW or scenario approaches, with $\theta^\ast = 0.9$ or $1.25 $, respectively.} 
    \label{fig:taxiNetExperiment}
\end{figure}

The aircraft starts at cross-track position $p_0 = 5$ meters, heading angle $\theta_0 = 10$ degrees, and runway down-track position of $322$ meters. 
At $422$ meters, wish to validate whether the cross-track position is within 1 meter of the centerline of the runway,
\begin{equation}
    C = \{p\in\mathbb{R}: |p| \leq 1\}.
\end{equation}
The SDF is,
\begin{equation}
    \label{eq:taxinetSpecification}
    g_{C}(p) = |p|-1,
\end{equation}
where we have taken the one norm deviation from the aircraft's crosstrack position.
Figure~\ref{fig:taxiNetExperiment} visually overviews the experiment and specification, \eqref{eq:taxinetSpecification}, in orange.
After we solve the root finding problem that attempts to compute the quantile in \eqref{eq:quantileCDF}, solving Problem~\ref{prob:setEnlargment} to find \eqref{eq:dkwEnlargment}.

We require that the specification in \eqref{eq:taxinetSpecification} be satisfied to with $1-\Delta=0.9$ probability. 
To determine if the specification is satisfied, we specify $\epsilon=0.1$ and $\beta=0.001$ for \eqref{eq:sampleReqDKW}, resulting in requiring 381 samples for the DKW-based approach. 
For the scenario approach, we specify a confidence of $\beta = 0.001$ to \eqref{eq:scenarioSamples}, requiring $126$ samples. 
We therefore ran the simulator for 381 times over a span of two hours. 
This resulted in determining that, by Problem~\ref{prob:NNProbVerification}, we did not satisfy the specification in \eqref{eq:taxinetSpecification}. 
Thus, we solve Problem~\ref{prob:setEnlargment}, to determine a $\theta$ that satisfies the probability of satisfaction $1-\Delta$. 
Therefore, we proceed to solve \eqref{opt:levelSet} in Problem~\ref{prob:setEnlargment} with the DKW and scenario methods, resulting in $\theta^*_{\rm DKW} = 0.9$ and $\theta^*_{\rm Scenario} = 1.25$, respectively.


\section{Conclusion}

In this paper, we have presented SAVER, a sampling-based toolbox for probabilistic verification of neural networks. 
The toolbox provides two approaches to obtain the number of samples necessary to verify whether we satisfy a specification. 
In addition, the toolbox is also able to modify the set through the usage of SDFs in order to achieve tha satisfaction probability.
To show the efficacy of this toolbox, we have presented its usage in three examples with different objectives. 

\bibliographystyle{IEEEtran}
\bibliography{refs}

\begin{thebibliography}{10}
\providecommand{\url}[1]{#1}
\csname url@samestyle\endcsname
\providecommand{\newblock}{\relax}
\providecommand{\bibinfo}[2]{#2}
\providecommand{\BIBentrySTDinterwordspacing}{\spaceskip=0pt\relax}
\providecommand{\BIBentryALTinterwordstretchfactor}{4}
\providecommand{\BIBentryALTinterwordspacing}{\spaceskip=\fontdimen2\font plus
\BIBentryALTinterwordstretchfactor\fontdimen3\font minus \fontdimen4\font\relax}
\providecommand{\BIBforeignlanguage}[2]{{%
\expandafter\ifx\csname l@#1\endcsname\relax
\typeout{** WARNING: IEEEtran.bst: No hyphenation pattern has been}%
\typeout{** loaded for the language `#1'. Using the pattern for}%
\typeout{** the default language instead.}%
\else
\language=\csname l@#1\endcsname
\fi
#2}}
\providecommand{\BIBdecl}{\relax}
\BIBdecl

\bibitem{neural_lander}
G.~Shi, X.~Shi, M.~O’Connell, R.~Yu, K.~Azizzadenesheli, A.~Anandkumar, Y.~Yue, and S.-J. Chung, ``Neural lander: Stable drone landing control using learned dynamics,'' in \emph{International Conference on Robotics and Automation (ICRA)}, 2019, pp. 9784--9790.

\bibitem{kaufmann2023champion}
E.~Kaufmann, L.~Bauersfeld, A.~Loquercio, , M.~Müller, V.~Koltun, and D.~Scaramuzza, ``Champion-level drone racing using deep reinforcement learning,'' \emph{Nature}, vol. 620, pp. 982--987, 2023.

\bibitem{autonomous_driving}
Y.~Huang and Y.~Chen, ``Survey of state-of-art autonomous driving technologies with deep learning,'' in \emph{IEEE 20th International Conference on Software Quality, Reliability and Security Companion (QRS-C)}, 2020, pp. 221--228.

\bibitem{ACASX}
K.~D. Julian, J.~Lopez, J.~S. Brush, M.~P. Owen, and M.~J. Kochenderfer, ``Policy compression for aircraft collision avoidance systems,'' in \emph{IEEE/AIAA 35th Digital Avionics Systems Conference (DASC)}, 2016, pp. 1--10.

\bibitem{taxinet}
D.~Cofer, I.~Amundson, R.~Sattigeri, A.~Passi, C.~Boggs, E.~Smith, L.~Gilham, T.~Byun, and S.~Rayadurgam, ``Run-time assurance for learning-based aircraft taxiing,'' in \emph{AIAA/IEEE 39th Digital Avionics Systems Conference (DASC)}, 2020, pp. 1--9.

\bibitem{malware_detection_attacks}
Q.~Wang, W.~Guo, K.~Zhang, A.~G. Ororbia, X.~Xing, X.~Liu, and C.~L. Giles, ``Adversary resistant deep neural networks with an application to malware detection,'' in \emph{Proceedings of the 23rd ACM SIGKDD International Conference on Knowledge Discovery and Data Mining}, New York, NY, USA, 2017, p. 1145–1153.

\bibitem{sparse_regression_attacks}
P.-Y. Chen, B.~Vinzamuri, and S.~Liu, ``Is ordered weighted $l_1$ regularized regression robust to adversarial perturbation? a case study on oscar,'' in \emph{2018 IEEE Global Conference on Signal and Information Processing (GlobalSIP)}, 2018, pp. 1174--1178.

\bibitem{MILP1}
A.~Lomuscio and L.~Maganti, ``An approach to reachability analysis for feed-forward relu neural networks,'' 2017.

\bibitem{MILP2}
O.~Bastani, Y.~Ioannou, L.~Lampropoulos, D.~Vytiniotis, A.~V. Nori, and A.~Criminisi, ``Measuring neural net robustness with constraints,'' in \emph{Proceedings of the 30th International Conference on Neural Information Processing Systems (NeurIPS)}.\hskip 1em plus 0.5em minus 0.4em\relax Red Hook, NY, USA: Curran Associates Inc., 2016, p. 2621–2629.

\bibitem{MILP3}
C.-H. Cheng, G.~N{\"u}hrenberg, and H.~Ruess, ``Maximum resilience of artificial neural networks,'' in \emph{Automated Technology for Verification and Analysis}.\hskip 1em plus 0.5em minus 0.4em\relax Cham: Springer International Publishing, 2017, pp. 251--268.

\bibitem{MILP4}
V.~Tjeng, K.~Y. Xiao, and R.~Tedrake, ``Evaluating robustness of neural networks with mixed integer programming,'' in \emph{International Conference on Learning Representations (ICLR)}, 2019.

\bibitem{SMT1}
L.~Pulina and A.~Tacchella, ``Challenging {SMT} solvers to verify neural networks,'' \emph{AI Communications}, vol.~25, no.~2, p. 117–135, Apr. 2012.

\bibitem{SMT2}
G.~Katz, C.~Barrett, D.~L. Dill, K.~Julian, and M.~J. Kochenderfer, ``Reluplex: An efficient {SMT} solver for verifying deep neural networks,'' in \emph{Computer Aided Verification}, R.~Majumdar and V.~Kun{\v{c}}ak, Eds.\hskip 1em plus 0.5em minus 0.4em\relax Cham: Springer International Publishing, 2017, pp. 97--117.

\bibitem{SDP_exact1}
R.~A. Brown, E.~Schmerling, N.~Azizan, and M.~Pavone, ``A unified view of {SDP}-based neural network verification through completely positive programming,'' in \emph{Proceedings of The 25th International Conference on Artificial Intelligence and Statistics}, ser. Proceedings of Machine Learning Research, vol. 151.\hskip 1em plus 0.5em minus 0.4em\relax PMLR, 28--30 Mar 2022, pp. 9334--9355.

\bibitem{SDP_exact2}
M.~Fazlyab, M.~Morari, and G.~J. Pappas, ``Safety verification and robustness analysis of neural networks via quadratic constraints and semidefinite programming,'' \emph{IEEE Transactions on Automatic Control}, vol.~67, no.~1, pp. 1--15, 2022.

\bibitem{SDP_exact3}
K.~D. Dvijotham, R.~Stanforth, S.~Gowal, C.~Qin, S.~De, and P.~Kohli, ``Efficient neural network verification with exactness characterization,'' in \emph{Proceedings of The 35th Uncertainty in Artificial Intelligence Conference}, ser. Proceedings of Machine Learning Research, vol. 115.\hskip 1em plus 0.5em minus 0.4em\relax PMLR, 22--25 Jul 2020, pp. 497--507.

\bibitem{SDP_exact4}
S.~Dathathri, K.~Dvijotham, A.~Kurakin, A.~Raghunathan, J.~Uesato, R.~R. Bunel, S.~Shankar, J.~Steinhardt, I.~Goodfellow, P.~S. Liang, and P.~Kohli, ``Enabling certification of verification-agnostic networks via memory-efficient semidefinite programming,'' in \emph{Advances in Neural Information Processing Systems}, vol.~33, 2020, pp. 5318--5331.

\bibitem{reachability}
J.~A. Vincent and M.~Schwager, ``Reachable polyhedral marching {(RPM)}: A safety verification algorithm for robotic systems with deep neural network components,'' in \emph{IEEE International Conference on Robotics and Automation (ICRA)}, 2021, p. 9029–9035.

\bibitem{random_attacks1}
S.-M. Moosavi-Dezfooli, A.~Fawzi, O.~Fawzi, and P.~Frossard, ``Universal adversarial perturbations,'' in \emph{2017 IEEE Conference on Computer Vision and Pattern Recognition (CVPR)}, 2017, pp. 86--94.

\bibitem{random_attacks2}
J.~Su, D.~V. Vargas, and K.~Sakurai, ``One pixel attack for fooling deep neural networks,'' \emph{IEEE Transactions on Evolutionary Computation}, vol.~23, no.~5, pp. 828--841, 2019.

\bibitem{PV1}
M.~Fazlyab, M.~Morari, and G.~J. Pappas, ``Probabilistic verification and reachability analysis of neural networks via semidefinite programming,'' in \emph{2019 IEEE 58th Conference on Decision and Control (CDC)}, 2019, pp. 2726--2731.

\bibitem{PV2}
J.~Pilipovsky, V.~Sivaramakrishnan, M.~Oishi, and P.~Tsiotras, ``Probabilistic verification of relu neural networks via characteristic functions,'' in \emph{Proceedings of The 5th Annual Learning for Dynamics and Control Conference}, ser. Proceedings of Machine Learning Research, vol. 211.\hskip 1em plus 0.5em minus 0.4em\relax PMLR, 15--16 Jun 2023, pp. 966--979.

\bibitem{PV3}
L.~Weng, P.-Y. Chen, L.~Nguyen, M.~Squillante, A.~Boopathy, I.~Oseledets, and L.~Daniel, ``{PROVEN}: Verifying robustness of neural networks with a probabilistic approach,'' in \emph{Proceedings of the 36th International Conference on Machine Learning}, ser. Proceedings of Machine Learning Research, vol.~97.\hskip 1em plus 0.5em minus 0.4em\relax PMLR, 09--15 Jun 2019, pp. 6727--6736.

\bibitem{PV4}
M.~Pautov, N.~Tursynbek, M.~Munkhoeva, N.~Muravev, A.~Petiushko, and I.~Oseledets, ``{CC-CERT}: A probabilistic approach to certify general robustness of neural networks,'' \emph{Proceedings of the AAAI Conference on Artificial Intelligence}, vol.~36, pp. 7975--7983, 06 2022.

\bibitem{branch_and_bound}
D.~Boetius, S.~Leue, and T.~Sutter, ``Probabilistic verification of neural networks using branch and bound,'' 2024.

\bibitem{scenario1}
B.~G. Anderson and S.~Sojoudi, ``Data-driven certification of neural networks with random input noise,'' \emph{IEEE Transactions on Control of Network Systems}, vol.~10, no.~1, pp. 249--260, 2023.

\bibitem{scenario2}
A.~Devonport and M.~Arcak, ``Estimating reachable sets with scenario optimization,'' in \emph{Proceedings of the 2nd Conference on Learning for Dynamics and Control}, ser. Proceedings of Machine Learning Research, vol. 120.\hskip 1em plus 0.5em minus 0.4em\relax PMLR, 10--11 Jun 2020, pp. 75--84.

\bibitem{dvoretzky1956asymptotic}
A.~Dvoretzky, J.~Kiefer, and J.~Wolfowitz, ``Asymptotic minimax character of the sample distribution function and of the classical multinomial estimator,'' \emph{The Annals of Mathematical Statistics}, pp. 642--669, 1956.

\bibitem{pmlr-v120-devonport20a}
A.~Devonport and M.~Arcak, ``Estimating reachable sets with scenario optimization,'' in \emph{Proceedings of the 2nd Conference on Learning for Dynamics and Control}, ser. Proceedings of Machine Learning Research, A.~M. Bayen, A.~Jadbabaie, G.~Pappas, P.~A. Parrilo, B.~Recht, C.~Tomlin, and M.~Zeilinger, Eds., vol. 120.\hskip 1em plus 0.5em minus 0.4em\relax PMLR, 10--11 Jun 2020, pp. 75--84.

\bibitem{tensorflow2015-whitepaper}
\BIBentryALTinterwordspacing
M.~Abadi~et al., ``{TensorFlow}: Large-scale machine learning on heterogeneous systems,'' 2015, software available from tensorflow.org. [Online]. Available: \url{https://www.tensorflow.org/}
\BIBentrySTDinterwordspacing

\bibitem{NEURIPS2019_9015}
\BIBentryALTinterwordspacing
A.~Paszke~et al., ``{PyTorch:} an imperative style, high-performance deep learning library,'' in \emph{Advances in Neural Information Processing Systems 32}.\hskip 1em plus 0.5em minus 0.4em\relax Curran Associates, Inc., 2019, pp. 8024--8035. [Online]. Available: \url{http://papers.neurips.cc/paper/9015-pytorch-an-imperative-style-high-performance-deep-learning-library.pdf}
\BIBentrySTDinterwordspacing

\bibitem{jax2018github}
\BIBentryALTinterwordspacing
J.~B. et~al., ``{JAX}: composable transformations of python numpy programs,'' 2018. [Online]. Available: \url{http://github.com/jax-ml/jax}
\BIBentrySTDinterwordspacing

\bibitem{harris2020array}
\BIBentryALTinterwordspacing
C.~R. Harris~et al., ``Array programming with {NumPy},'' \emph{Nature}, vol. 585, no. 7825, pp. 357--362, Sep. 2020. [Online]. Available: \url{https://doi.org/10.1038/s41586-020-2649-2}
\BIBentrySTDinterwordspacing

\bibitem{deng2012mnist}
L.~Deng, ``The {MNIST} database of handwritten digit images for machine learning research,'' \emph{IEEE Signal Processing Magazine}, vol.~29, no.~6, pp. 141--142, 2012.

\bibitem{Katz2021}
\BIBentryALTinterwordspacing
S.~M. Katz, A.~Corso, S.~Chinchali, A.~Elhafsi, A.~Sharma, M.~J. Kochenderfer, and M.~Pavone, ``{NASA ULI} aircraft taxi dataset,'' 2021. [Online]. Available: \url{https://purl.stanford.edu/zz143mb4347}
\BIBentrySTDinterwordspacing

\end{thebibliography}
\end{document}